\documentclass[letterpaper]{article}
\usepackage{iccc}

\usepackage{graphicx}
\usepackage{times}
\usepackage{helvet}
\usepackage{courier}
\title{Modern French Poetry Generation with RoBERTa and GPT-2}
\author{Mika Hämäläinen\textsuperscript{1}, Khalid Alnajjar\textsuperscript{1} and Thierry Poibeau\textsuperscript{2}\\
\textsuperscript{1}University of Helsinki, Finland\\
\textsuperscript{2}École Normale Supérieure-PSL and CNRS and Université Sorbonne nouvelle, Paris, France\\
firstname.lastname@\{helsinki.fi\textsuperscript{1} or ens.psl.eu\textsuperscript{2}\}\\
}
\begin{document} 
\maketitle
\begin{abstract}
\begin{quote}
We present a novel neural model for modern poetry generation in French. The model consists of two pretrained neural models that are fine-tuned for the poem generation task. The encoder of the model is a RoBERTa based one while the decoder is based on GPT-2. This way the model can benefit from the superior natural language understanding performance of RoBERTa and the good natural language generation performance of GPT-2. Our evaluation shows that the model can create French poetry successfully. On a 5 point scale, the lowest score of 3.57 was given by human judges to \textit{typicality} and \textit{emotionality} of the output poetry while the best score of 3.79 was given to \textit{understandability}.
\end{quote}
\end{abstract}

\section{Introduction}

Poem generation is a challenging creative natural language generation task. As a form of art, it has undergone several changes in the history. Classical poetry incorporates typically meter and rhyme as their function was to help people recall poems, especially when poetic tradition was still mostly oral rather than written.

In the modern era, the role of the poetry has changed from an art form that has to follow a fixed structure that defines its meter and rhyming such as iamb, haiku or anapest. Modern poetry is more concerned about creating something new by breaking any strict structural rules and by continuously questioning what poetry is, what it can be and what it should be (see \citeauthor{runoousoppi} 1990).

In the field of poem generation, meter is a feature that is very often considered in generated poetry \cite{colton2012full,lau2018deep,hamalainen2019let,zugarini2019neural,lewis2021syllable}. By incorporating meter, people can be more forgiving when evaluating the output of the system as it is known that people are ready to interpret more into the content of the output of a computationally creative system if the form is correct \cite{veale2016shape}. In other words, a poem that looks like a poem, as in that it follows a certain meter, must be a poem. A truly competent computational poet should be capable of generating something that is recognizable as a poem even if its output was modern free-form poetry.

In this paper, we explore the topic of modern poetry generation in French. We fine-tune a novel encoder-decoder architecture which consists of a RoBERTa \cite{liu2019roberta} based model as the encoder and a GPT-2 \cite{radford2019language} based model as the decoder. Because RoBERTa is very good at natural language understanding tasks but poor at generation tasks and GPT-2 is good at generation, but bad at understanding, it makes sense to incorporate both of the models. The task of RoBERTa is to encode the input (i.e. to understand poetry) and the task of GPT-2 is to decode the output (i.e. to generate poetry).

\section{Related work}

Poem generation has sparked a lot of interest in the past as we can see in a recent survey on the field \cite{goncalo-oliveira-2017-survey}. There is also some work on generating French poetry in particular \cite{van-de-cruys-2019-la,poibeau-etal-2020-sonnet}. In this section, we will provide a quick overview on some of the related work.

Poetry has been generated by using rules. This can, for instance, be done by modeling certain poetic features to find suitable verses \cite{twitsong3}, by writing different rules to generate verses with different poetic features \cite{hamalainen2018harnessing} or by using templates \cite{colton2012full}.

Genetic algorithms have been popular in poem generation as well \cite{manurung2012using,yang2016research,yang2021elitist}. In many of these approaches, the generation itself is not governed by rules, but rules are used to assess the quality of each generated poem. These approaches can be seen as a step away from fully relying on rules in the generative process.

Recently, different neural approaches have become common place in poetry generation such as LSTM models \cite{zhang2020automatic,atassi2022comparison} and GPT-2 \cite{hu2020generating,icaart22}. Our approach follows this line of work. How we set ourselves apart from the existing models is that we propose a novel poem generator model that can benefit from both RoBERTa and GPT-2 instead of relying only on one model. In addition, we condition the beam search on rhyme instead of solely relying on the output of a vanilla model.

\begin{table*}[]
\centering
\resizebox{\textwidth}{!}{%
\begin{tabular}{|l|l|l|l|}
\hline
Verse & Input                                       & Output                                             & Output in English                                         \\ \hline
1     & amant galamment escrime dame                & Vainement, paladin des dames, tu t'escrimes        & In vain, paladin of the ladies, you fight me with a sword \\ \hline
2     & Vainement, paladin des dames, tu t'escrimes & À crier que j'ai fait, ici, peu galamment.         & Shouting what I have done, here, little gallantly.        \\ \hline
3     & À crier que j'ai fait, ici, peu galamment.  & si de Laure Pétrarque avait été l'amant,           & if Pétrarque had been the lover of Laure                  \\ \hline
4     & si de Laure Pétrarque avait été l'amant,    & Qui sait ce qu'il eût dit de Laure dans ses Rimes? & Who knows what he said about Laure in his Rhymes?         \\ \hline
\end{tabular}%
}
\caption{Example of the training data for one poem}
\label{tab:train_example}
\end{table*}

\section{Data}

As machine learning requires data, we need a poem corpus. For this reason, we crawl all the French poems that are available on Wikisource\footnote{https://fr.wikisource.org/wiki/Catégorie:Poèmes}. The poems are not free of noise as some of the poems include verses in Greek alphabets, multiple different types of quotation marks, hyphens and spaces of different lengths etc. We clean the data from all of these inconsistencies by manually inspecting odd characters and either by replacing them (e.g. only one type of a hyphen) or removing them (e.g. Greek letters). The corpus contains 7553 poems. In addition, we use the French sonnet corpus introduced by \citeauthor{poibeau-etal-2020-sonnet} 2020. This corpus has 1039 sonnets.

Because these poems and sonnets are of different lengths, we split all of them into stanzas. From this point on, we treat a stanza as a poem so that all poems in our corpus are of a similar length. This gives us altogether 25,215 French poems and sonnets. For the purposes of our models, we do not make a distinction between poems and sonnets.

\section{Poem generator}

In this section, we describe our poem generation model. The model follows an encoder-decoder architecture where the encoder is a RoBERTa model and the decoder is a GPT-2 model. Rather than training these models from scratch, we use pretrained language models and fine-tune them for the task of poem generation using a transfer learning approach. We chose a RoBERTa-based model as the encoder given their great ability in capturing contextual semantics. GPT-2 is well-known for modeling a language; hence, making an optimal decoder for text-generation tasks.

First we have to pick the suitable pretrained models. As we use Transformers library \cite{wolf2020transformers}, we select our models from their repository. The current state-of-the-art French RoBERTa model is  CamemBERT\footnote{https://huggingface.co/camembert-base} \cite{martin-etal-2020-camembert} which is based on the RoBERTa \cite{liu2019roberta} architecture and trained on the large OSCAR corpus \cite{2022arXiv220106642A} in French. We use CamemBERT as our encoder.

As for the selection of the GPT-2 model, there were several alternatives. By trying the models out, we could see that all of them except for Belgian GPT-2\footnote{https://huggingface.co/antoiloui/belgpt2} \cite{louis2020belgpt2} predicted rather poor output. The model was trained on a variety of genres (such as news, Wikipedia, novels, European parliament text etc.) on a relatively big, around 60 GB, corpus. For this reason, we opted for Belgian GPT-2 as our decoder model.

We use Spacy\footnote{The \textit{fr\_core\_news\_sm} model} \cite{honnibal2020spacy} to extract up to 4 keywords from each poem in the corpus. We train our encoder-decoder architecture for sequence to sequence generation, where it predicts the next verse in a poem given a previous verse. In the absence of a previous verse, we train the model to predict the first verse of a poem from the up to 4 keywords extracted from the poem. An example of input and output in the training data for one poem can be seen in Table \ref{tab:train_example}. The first input consists of the keywords \textit{amant} (lover), \textit{galamment} (gallantly), \textit{escrime} (fencing) and \textit{dame} (lady), which are used to predict the first verse of the poem.

The poem corpus is split randomly to 80\% for training and 20\% for validation. The model is trained for 10 epochs. We use the Adam algorithm~\cite{Adam} with decoupled weight decay regularization~\cite{AdamW} and learning rate of 5e-05 to optimize the parameters of the model, with cross entropy loss as the loss function to reduce the difference between gold standard token and predicted tokens. 

Rhyming is taken into account during the generation phase. The model is requested to generate a sequence between 4 to 20 tokens, with a length penalty of 1.0 using a greedy approach. At each step of generating the output sequence (i.e., when predicting the next token), we use the model to predict the top 10 possible tokens instead of just one highest scoring output. We then sort these candidate tokens based on their probabilities and rhyming scores. The rhyming score is calculated by counting the number of tokens in the output that rhyme (full rhyme, consonance or assonance) with the input (i.e., the previous verse and any subsequent words generated during the run).

Because it is not easy to know whether two French words rhyme or not based on the orthography (similarly to English), we use eSpeak-ng\footnote{https://github.com/espeak-ng/espeak-ng} to produce an IPA (international phonetic alphabet) representation for each token in the model's vocabulary. IPA alphabets are designed to represent how words are pronounced by writing out the actual phonemes. We use a simple set of rules to compare the IPA strings of two tokens with each other to determine whether they rhyme or not. 

\begin{table*}[]
\centering
\begin{tabular}{ll}
\hline
\textit{\begin{tabular}[c]{@{}l@{}}D'un beau travail, d'une bonne pose,\\ De la paix, de la beauté.\\ Que je plains la beauté\\ De la femme, qui m'inspire\end{tabular}}                                   & \begin{tabular}[c]{@{}l@{}}From a beautiful work, from a good pose.\\ From the peace, from the beauty\\ Oh, I lament the beauty\\ Of the woman, who inspires me\end{tabular}                               \\ \hline
\textit{\begin{tabular}[c]{@{}l@{}}C'est ici que s'éveille le soleil,\\ C'est ici que repose le grand créateur,\\ Dont la ruine, hélas! se renouvelle\\ De l'Enfant du Progrès\end{tabular}}               & \begin{tabular}[c]{@{}l@{}}It is here where the sun wakes\\ It is here where the great creator rests\\ Whose ruin, alas! renews itself\\ From the Child of Progress\end{tabular}                         \\ \hline
\textit{\begin{tabular}[c]{@{}l@{}}C'est un des mois les plus beaux de l'année,\\ C'est le printemps, c'est l'été, c’est\\ Le ciel où mon printemps se joue.\\ À mon jardin qui s'effondrit.\end{tabular}} & \begin{tabular}[c]{@{}l@{}}It is one of the most beautiful months of the year,\\ It is the spring, it is the summer, it is\\ The sky where my spring plays.\\ In my garden that collapses\end{tabular} \\ \hline
\end{tabular}
\caption{Examples of generated poetry and their translations.}
\label{tab:examples}
\end{table*}

In practice, we first ensure that both of the IPA strings are equally long, if this is not the case, we remove characters from the beginning of the longer string until the IPA strings are equally long. If the strings are identical, no rhyme is considered, because a word does not make a good rhyme with itself. For full rhyme, the two IPA strings rhyme if they are identical from the first vowel onward. For assonance, we replace all consonants with a placeholder character \textit{C}, if the IPA strings are identical, i.e. they share the same vowels in the same positions, they are considered to have assonance rhyme. For consonance, we do the same as with assonance, but by replacing all vowels with a placeholder \textit{V}.

\section{Results and evaluation}

For evaluation purposes, we generate 20 different poems consisting of 4 verses each. For each poem, we use a set of four randomly selected keywords among all the keywords extracted from the poem corpus. None of the keyword combinations is identical to what the model saw during the training. We generate the poems similarly to the example shown in Table \ref{tab:train_example}. This means, that the keywords were used to generate the first verse, which was then used to generate the second verse and so on. 

Some of the generated poems and their translations can be seen in Table \ref{tab:examples}. As we can see, the generated output is cohesive and quite grammatical. We can, however, see that sometimes the verb conjugation might be wrong such as in the case of \textit{effondrit} which is a non-existing inflectional form of \textit{effondrer} (to collapse). Also, the model has a tendency of starting every verse with a capital letter even if it was a continuation to the sentence started in the previous verse.

We conduct a crowd-sourced evaluation on Appen\footnote{https://appen.com/}. We set French as a language requirement for the crowd-workers so that we know that they actually speak French and are able to assess French poetry. Each poem is evaluated by 20 different crowd-workers. An individual worker can evaluate all 20 different poems or just some of them, in which case the remaining unevaluated poems are shown to a different crowd-worker. An individual crowd-worker cannot evaluate the same poem multiple times.

For evaluation, we use the same parameters as used by several authors for evaluating poetry \cite{jukka,hamalainen-alnajjar-2019-generating,shihadeh2020emily}: (1) \textit{The poem is typical} (2) \textit{The poem is understandable} (3) \textit{The poem is grammatical} (4) \textit{The poem evokes imagination} (5) \textit{The poem evokes emotions} (6) \textit{I like the poem}. These statements are evaluated in a 5 point Likert scale, where 1 represents the worst and 5 the best grade.

\begin{table}[]
\centering
\begin{tabular}{|l|l|l|l|l|l|l|}
\hline
    & Q1   & Q2   & Q3    & Q4   & Q5   & Q6   \\ \hline
Avg & 3.57 & 3.79 & 3.77 & 3.65 & 3.57 & 3.77 \\ \hline
STD & 0.88 & 0.84 & 0.81  & 0.79 & 0.88 & 0.77 \\ \hline
\end{tabular}
\caption{The evaluation results and standard deviation}
\label{tab:evaluation_results}
\end{table}

The results can be seen in Table \ref{tab:evaluation_results}. All in all, the results are good and show that the system can generate poetry successfully. The lowest scores were obtained for \textit{typicality} and \textit{emotionality}. The highest score was given to \textit{understandability}. In the future, more robust human evaluation methods need to be applied to understand why these parameters scored high and low \cite{hamalainen-alnajjar-2021-great,hamalainen-alnajjar-2021-human}.

\section{Conclusions}

In this paper, we have presented a novel approach to French poem generation. We have presented an architecture that consists of RoBERTa and GPT-2 models that are fine-tuned on a poem corpus. In addition, we have modeled rhyme as a part of the prediction pipeline of the model.

The results obtained in human evaluation are promising and they indicate that the model performs well in the task it was designed to do. In order to make the evaluation results more transparent, we have released them in full on Zenodo\footnote{https://zenodo.org/record/6558357} together with the generated poems that were used in the evaluation.

Pretrained neural language models have been proven to be useful in poem generation. In the future, it would be interesting to study them in a multilingual setting, where a pretrained multilingual model is fine-tuned to generate poetry using the corpora of some languages other than the desired target language.

\section{Author Contributions}

The first two authors contributed to the work presented in this paper equally. The third author was involved in planning the methods and writing the paper.

\section{Acknowledgments}

This work was partially financed by the Society of Swedish Literature in Finland with funding from Enhancing Conversational AI with Computational Creativity. This work was conducted during a mobility period supported by Nokia Foundation under grant number 20220193. This work was supported in part by the French government under management of Agence Nationale de la Recherche as part of the “Investissements d’avenir” program, reference ANR19-P3IA-0001 (PRAIRIE 3IA Institute). The work was also supported by the CNRS funded International Research Network Cyclades (Corpora and Computational Linguistics for Digital Humanities).






\bibliographystyle{iccc}
\bibliography{iccc}

\end{document}